\documentclass{article}
\usepackage{iclr2026_conference,times}

\usepackage{amsmath,amsfonts,bm}

\def\eqref#1{equation~\ref{#1}}
\def\1{\bm{1}}

\DeclareMathAlphabet{\mathsfit}{\encodingdefault}{\sfdefault}{m}{sl}
\SetMathAlphabet{\mathsfit}{bold}{\encodingdefault}{\sfdefault}{bx}{n}

\usepackage{hyperref}
\usepackage{url}
\usepackage[utf8]{inputenc}
\usepackage[T1]{fontenc}
\usepackage{booktabs}
\usepackage{bookmark}
\usepackage{amsfonts}
\usepackage{nicefrac}
\usepackage{microtype}
\usepackage{xcolor}
\usepackage{multirow}
\usepackage{graphicx}
\usepackage{amsmath}
\usepackage{makecell}
\usepackage{xspace}
\usepackage{subcaption}
\usepackage{comment}
\usepackage{ulem}
\usepackage{xcolor}
\usepackage{fvextra}
\definecolor{embldarkGreen}{HTML}{0A5032}
\definecolor{emblGreen}{HTML}{18974C}

\newcommand{\librarian}{\textsc{EMBL AI Librarian}}
\newcommand{\libshort}{\textsc{Librarian}}

\title{\textcolor{emblGreen}{\textbf{\librarian{}:}}\\
Life-Sciences Knowledge Layer for AI Agents}
\author{Luigi~Sigillo$^{1}$\thanks{Corresponding Author} \quad
Matteo~Silvestri$^{1,3}$ \quad
Francesco~Tabaro$^{1}$ \quad
Rajat~Bhatnagar$^{2}$ \\
\bfseries
Syed~Irtaza~Mubashar$^{2}$ \quad
Matt~Jeffryes$^{2}$ \quad
Daljit~Nijjer$^{2}$ \quad
Vittorio~Perera$^{1}$\\
\bfseries
Ola~Spjuth$^{4,5}$ \quad
Julio~Saez-Rodriguez$^{2,6}$ \quad
Melissa~Harrison$^{2}$ \quad
Fabio~Petroni$^{1,2}$
\\[-0.5em]
\normalfont
\parbox{\textwidth}{\rule{0pt}{28pt}{\normalfont\noindent%
$^{1}$EMBL Rome, European Molecular Biology Laboratory, Monterotondo, Italy.
$^{2}$European Bioinformatics Institute (EMBL-EBI), European Molecular Biology Laboratory, Hinxton, UK. 
$^{3}$Sapienza University of Rome, Italy.
$^{4}$Department of Pharmaceutical Biosciences, Uppsala University, Uppsala, Sweden.
$^{5}$Science for Life Laboratory (SciLifeLab), Uppsala University, Uppsala, Sweden.
$^{6}$Institute for Computational Biomedicine, Heidelberg University, Germany.\\
\texttt{luigi.sigillo@embl.it}
}}
}
\iclrpreprint
\begin{document}

\maketitle

\begin{figure}[ht]
\vspace{-25pt}
    \centering
    \includegraphics[width=\linewidth]{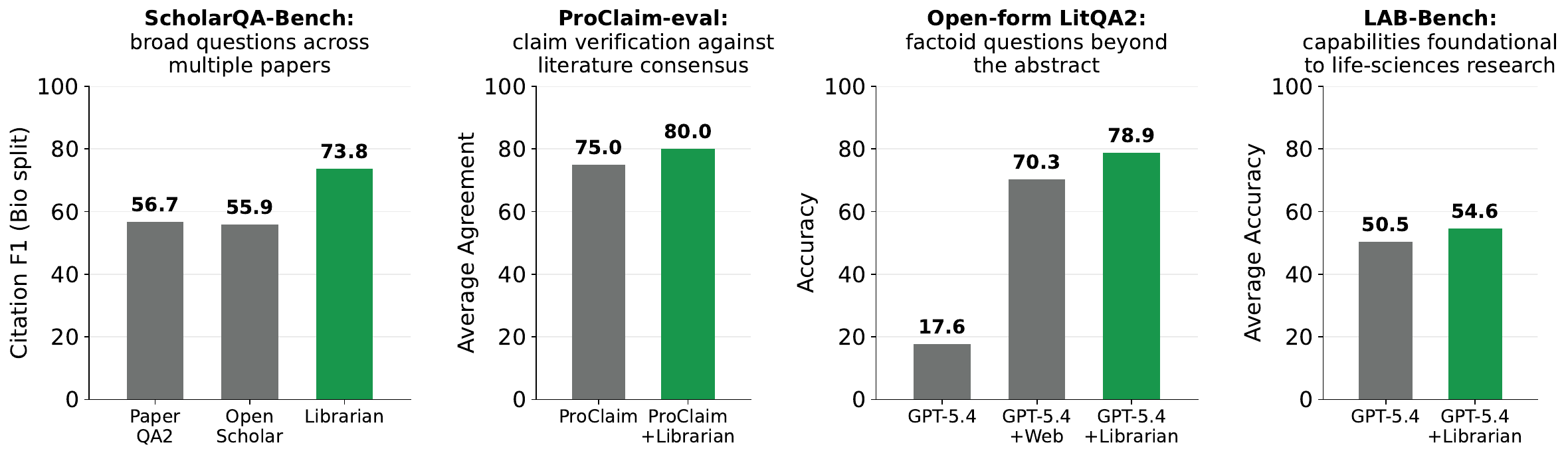}
    \vspace{-7mm}
    \caption{On four benchmark suites, agents equipped with the \libshort{} knowledge layer (green) outperform their baselines (grey): literature synthesis (ScholarQA-Bench, Citation F1 on the Bio split), claim verification (ProClaim-eval, average agreement with expert consensus), open-form question answering (LitQA2, accuracy), and foundational biology tasks (LAB-Bench, average accuracy).}
    \label{fig:first_image}
    \vspace{-4pt}
\end{figure}

\begin{abstract} 
The web is increasingly accessed by AI agents rather than humans.
Every agent needs knowledge, especially in the life-sciences, where agentic pipelines are growing fast. Access to the literature is a crucial part of that need, and resources such as Europe PMC, with over 40\,M indexed records, are widely used to meet it. Yet these resources were not built for AI agents: they take keywords and complex syntax and return whole papers, so every agent must learn the syntax, issue several searches, and read full papers to find the evidence it needs.
We introduce \librarian{}, a knowledge layer that upgrades the Europe PMC interface for AI agents: an agent asks in natural language and receives evidence that answers it. 
A single LLM orchestrates the whole knowledge retrieval process: it plans complementary subqueries executed by the live Europe PMC search engine, then reads the selected papers and locates the relevant evidence.
We evaluate \libshort{} across four benchmarks: literature synthesis, claim verification, open-domain question answering, and downstream biology tasks such as protocol questions and sequence manipulation. On ScholarQABench, \libshort{} improves Citation F1 by more than $16$ points over strong recently published baselines. Used as the retrieval layer of an existing claim-verification pipeline, it increases agreement with expert consensus; and on the open-form LitQA2 benchmark, a GPT-5.4 agent scores about $8$ points higher when grounded in \libshort{} than with web search. Overall, our results show that equipping life-science agents with the \libshort{} knowledge layer improves performance across a range of tasks.
We release our code publicly at \url{https://github.com/petroni-lab/librarian}.
\end{abstract}

\section{Introduction}
\label{sec:intro}  

Recent measurements indicate that traffic from automated agents, many powered by large language models (LLMs), now rivals or exceeds human-generated traffic across much of the web~\citep{imperva2025badbot, human2026_StateofAITraffic}, motivating a rethinking of web infrastructure toward an agent-first approach~\citep{bandara2026agentfirstwebredesigningweb, yang2025agenticwebweavingweb}. 
This rethinking is not confined to the consumer web. In the life sciences, agents are already embedded in everyday research workflows: they help researchers summarize findings across many papers~\citep{skarlinski2024language, Asai2026}, execute code and orchestrate bioinformatics tools~\citep{huang2025biomni, gao2025tooluniverse, Aygun2026}, prioritize candidate biomarkers~\citep{kim2026codasaicodatascientistbiomarker}, and generate testable hypotheses~\citep {mitchener2025kosmos, gottweis2025aicoscientist, ghareeb2026robin}.
Common to all of these applications is a dependence on the scientific literature: whether an agent is summarizing findings or proposing an hypothesis, it must anchor its reasoning in published evidence. Providing that evidence, however, is far from straightforward: the literature is growing exponentially~\citep{Bornmann2021}, intensifying the information overload in research~\citep{klerings2015information, gusenbauer2021age, galli2025}. 

Europe PMC is an open literature search engine for the life sciences~\citep{rosonovski2024,ferguson2021}, with coverage of medicine and health subjects that exceeds PubMed~\citep{gusenbauer2022searchwhereyouwillfindmost}. Its holdings include 11.9\,M full-text articles, 40.7\,M PubMed abstracts, and 1.2\,M preprint records.
The service, freely accessible via website, API, and bulk download, is hosted by EMBL-EBI and grounded in the open-access mandates of a consortium of international life-science funders~\citep{rosonovski2024}. 
Its interface exposes a rich query syntax over metadata and article content, including full-text sections such as methods and figures, and annotations for entities such as chemicals, organisms, gene/protein names, and diseases. These search capabilities have been progressively extended and refined through iterative, user-centred development informed by usability studies and community feedback \citep{levchenko2018europe, rosonovski2024}. Today Europe PMC serves a large international user community, receiving access from more than 32 million unique IP addresses in 2023 \citep{cordis2025}. 

Life-science agents are increasingly part of the Europe PMC user base~\citep{huang2025biomni, mitchener2025kosmos, skarlinski2024language}.
Yet that interface was built for humans, not for agents. On the input side, agents must use keyword-based queries rather than natural language, and because biological entities have multiple aliases (\textit{e.g.}, gene symbols, protein names, disease synonyms, or MeSH terms), multiple complementary searches are often required for comprehensive retrieval~\citep{wang2023can, wang2025reassessing}. On the output side, results are whole documents ranked by lexical score, of which only a fraction is relevant. 
Reading full papers is especially costly for agents with fixed context windows, as each document consumes a large fraction of the token budget~\citep{hsu2026rethinkingagenticsearchpiserini, hu2026sagebenchmarkingimprovingretrieval}.
These challenges motivate a novel life-sciences knowledge layer for AI agents: a shared component, queried in natural language, that returns a compact set of citable evidence snippets, which recent work suggests might improve performance more than scaling the model alone~\citep{nasri2026virbench}.

A common way to expose the literature to AI agents is a dense vector database: papers are chunked into passages, embedded into vectors, and the nearest passages are retrieved so an agent can query the database in natural language and receive citable passages rather than read whole papers~\citep{karpukhin2020dense, NEURIPS2020_6b493230, skarlinski2024language, volkova2025crossdisciplinaryknowledgeretrievalsynthesis, shi2025scisage, Asai2026, ding2026scirag}. However, a dense vector index has drawbacks. First, it is expensive: indexing and serving the whole literature demands a heavy infrastructure investment with hundreds of gigabytes of embeddings (OpenScholar's datastore alone measures roughly 744\,GB~\citep{Asai2026}). Second, the margin dense retrieval offers is narrowing as LLMs grow more capable: a well-tuned BM25 backbone driven by a capable LLM issuing multiple keyword queries matches or exceeds dense retrieval on scientific benchmarks~\citep{hsu2026rethinkingagenticsearchpiserini,
hu2026sagebenchmarkingimprovingretrieval}. Third, embeddings flatten structured metadata, preventing direct queries over explicit fields such as gene, protein, organism, and chemical annotations.
Finally, a nearest-neighbour lookup over vectors is not inspectable, unlike a structured, fielded keyword query.

We introduce \librarian{}, an agent-first knowledge layer on top of Europe PMC: it preserves the natural-language, citable-evidence interface of dense retrieval but maintains no index of its own, instead querying Europe PMC's live search directly. A single LLM controller generates complementary keyword and fielded queries, retrieves the matching records, decomposes
selected papers into paragraphs, scores candidate evidence against the original question, and returns a compact, ranked set of citable evidence with source metadata. \librarian{} is model-agnostic: any LLM can serve as its engine, as newer, more capable models emerge, the system inherits the benefit.

% 7 evaluation campaign and results
We evaluate \libshort{} across four settings that require grounding in the life-science literature: literature synthesis (writing evidence-backed summaries over multiple papers), claim verification (checking whether published evidence supports a scientific claim), open-domain question answering, and biology workflow tasks (\textit{e.g.}, protocol questions, sequence manipulation, and molecular cloning). In every setting, equipping agents with the \libshort{} knowledge layer yields improvements (Figure~\ref{fig:first_image}). For literature synthesis, it raises Citation F1 by more than $16$ points on average over strong recently published agentic baselines. For claim verification, using \libshort{} as the evidence-retrieval layer of the ProClaim pipeline~\citep{ai2026proclaim} increases agreement by $5$ points on average. For open-domain QA, equipping a frontier model (GPT-5.4) with \libshort{} improves LitQA2 accuracy by ${\sim}8$ points over web search~\citep{skarlinski2024language}. For biology workflow tasks (LAB-Bench~\citep{laurent2024lab}), \libshort{} improves macro accuracy for both frontier backbones (up to $+4.2$ points), with large gain of $+11.3$ on manipulating biological sequences.

% 8. contributions
To summarize, our contributions are as follows:
\begin{enumerate}
    \item  We define a life-science knowledge-layer interface for the agentic era: an agent sends a natural-language question and receives a compact set of citable evidence snippets, without learning a query syntax or reading whole papers.

    \item We introduce \librarian{}, a system architecture that implements this interface on top of Europe PMC, reusing its curated, fielded life-science search infrastructure. A single,
    model-agnostic LLM controller orchestrates the entire retrieval pipeline.

    \item We run an extensive evaluation across multiple datasets from four complementary benchmark suites, showing that equipping diverse agents with the same \libshort{} layer consistently improves their performance.

    \item We release the code, prompts, runtime configuration, and evaluation pipelines needed to reproduce our results at \url{https://github.com/petroni-lab/librarian}.
\end{enumerate}

\section{\librarian{}} 
\label{sec:method} 

\begin{figure*}[t]
\centering
    \includegraphics[width=1\linewidth]{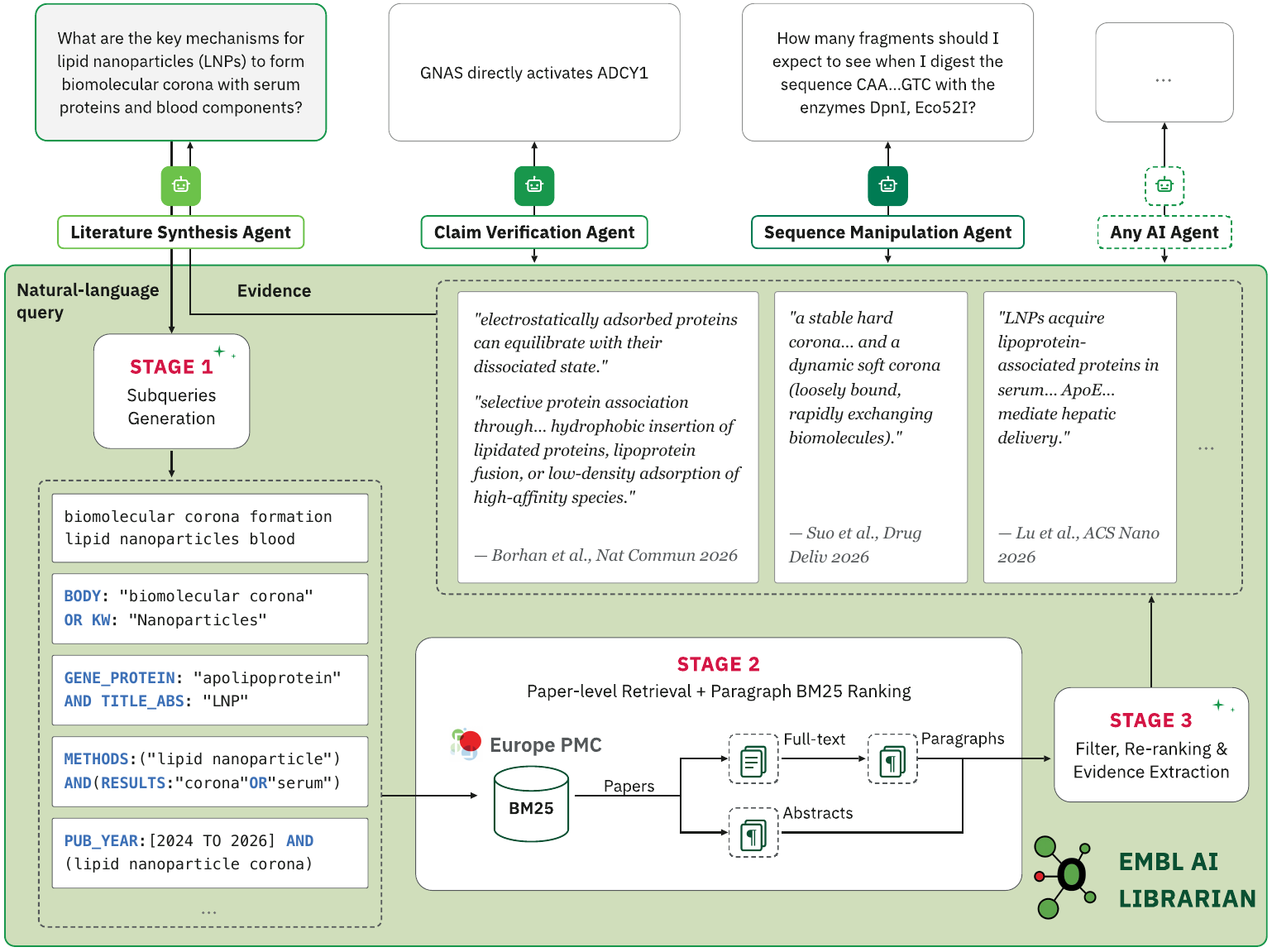}
\caption{Overview of \librarian{}. Starting from a natural-language query, \libshort{} runs three stages: (Stage~1) generates complementary subqueries in Europe PMC's advanced query syntax and validates them; (Stage~2) executes the validated subqueries against the live Europe PMC search engine, retrieves candidate papers, and ranks paragraphs from their full text; (Stage~3) filters and re-ranks the top-ranked paragraphs and abstracts, then extracts and returns the final evidence.}
\label{fig:arch}
\vspace{-0.8mm}
\end{figure*}

Given a user query $q$, \libshort{} returns a ranked list of evidence items $\mathcal{E}=[e_1,\ldots,e_m]$. We define an evidence item as the set comprising a short text span, paper metadata for citation, and the Europe PMC query used for retrieval (Figure~\ref{fig:arch}). This is a much finer-grained unit than a conventional search result, which in Europe PMC corresponds to entire articles. Because \libshort{} returns evidence rather than whole documents, downstream agents can directly quote, cite and reason over the results, without writing Europe PMC queries or reading full papers. At the same time, each item keeps the query that produced it, so retrieval stays transparent: an agent (or the user) can check how the evidence was found. We next describe how this evidence is retrieved and ranked. 

\subsection{Evidence retrieval} 
\librarian{} comprises three stages (Figure~\ref{fig:arch}). It begins with recall-oriented retrieval because life-science concepts are expressed through many surface forms, including gene symbols (\textit{e.g.}, TP53), protein names (\textit{e.g.}, p53), disease names (\textit{e.g.}, ALS or amyotrophic lateral sclerosis), organism constraints, and controlled vocabularies such as MeSH. A record indexed under one form may be missed by a Europe PMC query targeting another, limiting the recall of any single query. The subsequent stages rank evidence within retrieved papers and filter false positives, selecting the evidence most relevant to the original question. Overall, \libshort{} progresses from recall-oriented retrieval to precision-oriented evidence selection.

\textbf{Stage 1: Subquery generation.}
\libshort{} maps the user query $q$ to a set of complementary subqueries $Q = f_{\mathrm{plan}}(q) =\{q_1, \ldots, q_N\} ,$ where $f_{\mathrm{plan}}$ is an LLM. A single prompt converts the user query into a JSON list of subqueries; each $q_i$ is a keyword or fielded subquery that can target title, abstract, keyword/MeSH terms, gene or protein, organism, disease, chemical, and other Europe PMC fields (prompt and field grammar in Appendix~\ref{app:query-prompt}). Each subquery passes a deterministic syntax validator before execution, and malformed expressions are dropped; only subqueries that pass validation are executed as Europe PMC queries. The maximum number of subqueries $N$ is fixed.

\textbf{Stage 2: Paper-level retrieval and paragraph ranking.}
The validated subqueries are executed in parallel as Europe PMC queries against the live Europe PMC search service, which returns records in relevance order. Each Europe PMC query returns up to $P$ records, for a pool of at most $M = N \times P$ records; the concrete values are given in Section~\ref{sec:setup}. Complementary subqueries can return the same paper, so \libshort{} keeps only a single version of each. Each full-text paper is decomposed into paragraphs: \libshort{} retrieves the XML and extracts the body paragraphs, discarding references and other front/back matter. It ranks these paragraphs with BM25~\citep{robertson1994okapi}, taking their union as the corpus. Each paragraph is concatenated with its section title and type (\textit{e.g.}, methods, results, discussion), scored against the original question $q$. Each paper passes $k$ paragraphs to Step 3: always its abstract (\textit{i.e.}, $k\ge1$), plus the highest-ranked paragraphs when open-access full text is available. Papers without full text contribute only their abstract, which enters Stage 3 directly.

\textbf{Stage 3: Filtering, re-ranking, and evidence extraction.}
At this point, every candidate paper is represented as a list of paragraphs. \libshort{} concatenates these paragraphs, segments them into sentences with pySBD~\citep{pysbd}, and prepends a unique identifier to each sentence in the prompt. A single LLM call then evaluates a batch of candidate papers and returns an ordered list of sentence identifiers (prompt in Appendix~\ref{app:ft-filter-prompt}). This call serves three purposes: (i) filtering out irrelevant paragraphs, (ii) re-ranking relevant paragraphs, and (iii) extracting their supporting evidence as sentences. This is the precision-oriented step: the model cross-encodes the original question $q$ with every sentence in one pass, jointly attending over both for a fine-grained relevance assessment that lexical scoring cannot provide. Because the model emits short identifiers instead of regenerating text, it decodes far fewer tokens, lowering latency and cost.
Finally, \libshort{} assembles the output: it returns the selected sentences, clustered by source paper and paired with its metadata. The result $\mathcal{E}$ is a ranked set of citable evidence snippets that downstream agents can use for the task at hand (\textit{e.g.}, synthesis, claim verification, or question answering).

% ============================================================
%  EXPERIMENTS & RESULTS
% ============================================================

\section{Experiments}
\label{sec:setup}
We evaluate whether task-specific agents improve when equipped with the \libshort{} knowledge layer, across four benchmarks that probe complementary capabilities: citation-grounded synthesis across multiple papers on ScholarQA-Bench (Section~\ref{sec:oscholar_neu_bio}); claim verification against the literature consensus on ProClaim-Eval (Section~\ref{sec:proclaim}); factoid question answering beyond the abstract, on open-form LitQA2 (Section~\ref{sec:openform}); and capabilities foundational to life-sciences research, from database and sequence questions to cloning and protocols, on LAB-Bench (Section~\ref{sec:labbench}). Together they range from open-ended synthesis to narrow factoid retrieval, isolating where the knowledge layer helps most.

\textbf{Implementation Details.}
In our experiments, we use GLM-5~\citep{glm5team2026glm5vibecodingagentic} for \libshort{}. GLM-5 is self-hosted on an NVIDIA DGX B200 and served through vLLM~\citep{kwon2023efficient}. In our evaluation, \libshort{} uses $N=7$ subqueries and retrieves up to $P=50$ articles per subquery. For full-text articles, $k=16$ paragraphs reach Stage 3.

\subsection{ScholarQA-Bench: broad questions across multiple papers}
\label{sec:oscholar_neu_bio}

ScholarQABench~\citep{Asai2026} is a collection of broad questions written by PhD-level experts across multiple fields; answering each requires synthesizing evidence from several articles. We use its biomedicine (Bio) and neuroscience (Neu) splits, together with the subset of multidisciplinary (Multi) questions that fall under the Bio and Neu categories.

\textbf{Agents.} We build a \textit{Synthesis Agent} that, given a broad question, issues a retrieval call and generates a citation-grounded summary over the evidence. As retrievers, we consider \libshort{} and a BM25 index over the OpenScholar Data Store (OSDS). The agent is powered by GLM-5 (one prompt), and we compare it against the best-performing agents from \citet{Asai2026,ding2026scirag}.

\textbf{Metrics.} Following~\citet{Asai2026}, we report \textit{Citation F1} and \textit{LLM}. \textit{Citation F1} is the harmonic mean of citation precision and recall: recall measures whether every citation-worthy statement is supported by appropriate references, while precision measures whether each cited source is both relevant to and necessary for the statement it supports. \textit{LLM} is the average score (1--5) from an LLM-as-a-judge (\textit{i.e.}, Prometheus~\citep{kim2023prometheus}) across three dimensions: organization, coverage, and relevance. See~\citet{Asai2026} for details.

% =========================
\begin{table*}[t]
\centering
\caption{ScholarQABench results. As \textit{Retriever}, we consider OpenScholar, SciRAG, and BM25 over the OpenScholar Data Store (OSDS); PaperQA2 on its own corpus; and \libshort{} over Europe PMC. We report \textit{Citation F1} on the Bio, Neu, and Multi splits, and the \textit{LLM} score (1--5) on Multi.}
\label{tab:scholarqa_results}
\resizebox{\textwidth}{!}{%
\begin{tabular}{llcccc}
\toprule
& &  \multicolumn{3}{c}{\textbf{Citation F1}} & \multicolumn{1}{c}{\textbf{LLM}} \\
\cmidrule(lr){3-5} \cmidrule(lr){6-6}
\textbf{Agent} & \textbf{Retriever} & \textbf{Bio} & \textbf{Neu} & \textbf{Multi} & \textbf{Multi} \\
\midrule

%Llama-3.1-8B \citep{grattafiori2024llama} & OSDS & OpenScholar & 38.0 & 36.8 & -- & -- \\
OpenScholar-8B \citep{Asai2026} & OpenScholar on OSDS & 50.8 & 56.8 & 34.7 & \underline{4.18}  \\
%Llama-3.1-70B \citep{grattafiori2024llama} & OSDS & OpenScholar & 53.8 & 58.1 & -- & -- \\
OpenScholar-70B \citep{Asai2026} & OpenScholar on OSDS & 55.9 & 63.1 & -- & -- \\
%GPT-4o \citep{openai2024gpt4ocard} & -- & None & 0.2 & 0.1 & -- & -- \\
GPT-4o \citep{openai2024gpt4ocard} & OpenScholar on OSDS & 36.3 & 21.9 & 54.3 &3.79 \\
%OpenScholar-GPT-4o \citep{Asai2026} & OSDS & OpenScholar & 51.5 & 43.5 & -- & -- \\
PaperQA2 \citep{skarlinski2024language} &  PaperQA2 & 56.7 & 56.0 & -- & -- \\
SciRAG (Llama-3.1-70B) \citep{ding2026scirag} & SciRAG on OSDS   & 43.1 & 45.9 & -- & -- \\
SciRAG (GPT-4o) \citep{ding2026scirag} & SciRAG on OSDS & 44.8 & 36.2 & -- & -- \\
\midrule

% Synthesis Agent (GLM-5) \fabio{move to ablation} & EPMC & \libshort{} (Abstract) & 70.6 & 73.0 & -- & -- \\
% Synthesis A + \librarian{} (abstract-only, GLM-5) & OSDS & \libshort{} (BM25) & 71.2 & 73.9 & -- & -- \\

% \midrule

% Synthesis A + \librarian{} (hierarchical, Qwen3.5-27B) & OSDS & \libshort{} (Hierarchical) & 50.2 & 51.5 & 3.70 & 28.22 \\
% Synthesis A + \librarian{} (hierarchical, Qwen3.5-27B) & Europe PMC & \libshort{} (Hierarchical) & 50.2 & 51.5 & 3.50 & 42.84 \\
% Synthesis Agent (GLM-5) \fabio{move to ablation}  & EPMC & \libshort{} (Hierarchical) & \textbf{77.6} & 80.2 & 4.73 & \textbf{74.5} \\
% \textbf{Synthesis A + \librarian{} (hierarchical, GLM-5)} & OSDS & Librarian (Hierarchical) & 75.8 & \textbf{80.4} & \textbf{4.82} & 71.06 \\

%\midrule
Synthesis Agent (GLM-5) & BM25 on OSDS & \underline{67.0} & \underline{68.5} & \underline{72.0} & 4.10 \\
% Synthesis A + \libshort{} (bm25, Qwen3.5-27B) & Europe PMC & \libshort{} (BM25) & ?? & ?? & -- & -- \\
Synthesis Agent (GLM-5)  &  \libshort{} & \textbf{73.8} & \textbf{79.4} & \textbf{76.2} &\textbf{4.90}  \\
% Synthesis Agent (GPT-4o) & EPMC & \libshort{} (BM25) & -- & -- & 4.33 & 40.5* \\
% Synthesis Agent (GPT-5.4) & EPMC & \libshort{} (BM25) & -- & -- & 4.56 & 46.3* \\

\bottomrule
\end{tabular}
}

\end{table*}
% =========================

\textbf{Findings.}
Table~\ref{tab:scholarqa_results} reports our results. First, the Synthesis Agent outperforms the baselines despite using the same retrieval corpus (OSDS), improving \textit{Citation F1} by $+11.1$ on \textit{Bio} and $+5.4$ on \textit{Neu} over OpenScholar-70B. Since GLM-5 has an order of magnitude more parameters than the baseline agents ($\sim700B$ vs $\sim70B$), this suggests, as expected, that a larger model grounds its answers in retrieved evidence more effectively. The Synthesis Agent becomes even stronger when equipped with \libshort{}, with a further $+6.8$ on \textit{Bio} and $+10.9$ on \textit{Neu} over the BM25 index on OSDS. We attribute this to the \libshort{} pipeline providing better evidence, a hypothesis we further validate on other benchmarks below. On \textit{Multi}, the \textit{LLM} score of the Synthesis Agent with OSDS is on par with the baselines, suggesting that a larger model alone does not help here; the \libshort{} knowledge layer, in contrast, yields a $20\%$ improvement, further supporting our hypothesis.

\subsection{ProClaim-eval: claim verification against literature consensus}
\label{sec:proclaim}
ProClaim-eval~\citep{ai2026proclaim} evaluates an agent's ability to determine, from the literature, the scientific consensus on a given claim. The task requires retrieving evidence from multiple articles and assigning one of three verdicts: \textsc{Support} (the evidence supports the claim), \textsc{Refute} (the evidence refutes it), or \textsc{Uncertain} (the evidence is insufficient). The benchmark contains 419 claims from SIGNOR~\citep{loSurdo2025signor4} (protein--protein interactions) and ConnectomeDB~\citep{liu2026connectomedb2025} (ligand--receptor interactions), each with an expert-curated consensus verdict.

\textbf{Agents.}
The ProClaim agent~\citep{ai2026proclaim} orchestrates three stages inside a refinement loop: (i) \textit{retrieval} searches for evidence on the claim; (ii) a \textit{sub-agent} reads the retrieved full-text papers, extracts claim-relevant facts, and assigns each a verdict label; and (iii) a \textit{sufficiency classifier} decides whether the collected evidence suffices to reach a verdict. If not, the loop restarts from (i). As the retriever, we consider either the original from \citet{ai2026proclaim}, which combines PubMed and Semantic Scholar (S2), or \libshort{} over Europe PMC. Adapting ProClaim to \libshort{} requires minor modifications to stages (ii) and (iii). In (ii), fact extraction no longer mines full-text papers but directly labels the evidence \libshort{} returns. In (iii), ProClaim's MLP classifier relies on features (impact factor, citation counts, full-text NLI signals) that \libshort{} does not provide, so we remove them from the prompt. We also implement a simpler \textit{Verifier Agent} (a single prompt) that, given a claim and the retrieved evidence, emits a verdict. As a baseline, we include the OpenScholar agent~\citep{Asai2026} over OSDS. All agents are powered by Claude Sonnet 4.6~\cite{anthropic2026claudesonnet46}.

\textbf{Metrics.}
We report \textit{Agreement}, defined as the fraction of claims assigned the correct verdict;
higher values of AGR indicate stronger alignment between model predictions and gold annotations.

\begin{table*}[t]
\centering
\caption{
ProClaim-eval results. We evaluate the ProClaim agent~\citep{ai2026proclaim}, a simpler Verifier Agent, and an OpenScholar baseline, all powered by Claude Sonnet 4.6. The retrievers are OpenScholar over Semantic Scholar (S2), the ProClaim retriever (PubMed and S2), and \libshort{} over Europe PMC. Agreement with expert consensus is reported on average (AVG) and per dataset.
}
\label{tab:proclaim}
%\resizebox{\textwidth}{!}{
\begin{tabular}{ll c|cc}
\toprule
\multirow{2}{*}{\textbf{Agent}}
& \multirow{2}{*}{\textbf{Retriever}}
& \multicolumn{3}{c}{\textbf{Agreement\,$\uparrow$}}
\\
\cmidrule(l){3-5}
&
&
\textbf{AVG}
&
\textbf{SIGNOR}
&
\textbf{ConnectomeDB}
\\
\midrule
OpenScholar
& OpenScholar on S2
& 0.43
& 0.40
& 0.46 \\
Verifier Agent
& PubMed + S2 Search
& 0.51
& 0.62
& 0.48
\\
ProClaim
& PubMed + S2 Search
& \underline{0.75}
& \underline{0.75}
& \underline{0.75}
\\
\midrule
Verifier Agent
& \libshort{}
& 0.66
& 0.65
& 0.66 \\
ProClaim
& \libshort
& \textbf{0.80}
& \textbf{0.78}
& \textbf{0.81}
\\
\bottomrule
\end{tabular}
%}
\end{table*}

\textbf{Findings.}
Table~\ref{tab:proclaim} reports the results. First, the one-prompt Verifier Agent, when equipped with \libshort{}, is already strong: it reaches $0.66$ average agreement, $+15$ points over retrieving from PubMed and Semantic Scholar (S2) and only $9$ points below the original ProClaim pipeline. This suggests that \libshort{} already does much of the heavy lifting by surfacing the relevant evidence, leaving the downstream agent only the task of turning that evidence into a verdict. The same holds for ProClaim itself: switching its retriever to the \libshort{} knowledge layer yields $+5$ average agreement points ($0.75 \to 0.80$), further supporting the hypothesis that \libshort{} provides better access to the underlying knowledge. Both agents far exceed the OpenScholar baseline ($0.43$).

\subsection{Open-form LitQA2: factoid questions beyond the abstract}
\label{sec:openform}
LitQA2~\citep{skarlinski2024language} is a benchmark of narrow, difficult factoid scientific questions designed to be unanswerable from the abstract alone, requiring full-text evidence. It is distributed as a multiple-choice benchmark; we create an open-form version in which the answer must be produced directly from the literature rather than selected from candidate options. We retain the $91$ questions whose supporting paper is available as full text in Europe PMC.

\textbf{Agents.}
We consider a RAG-style \textit{QA Agent} \citep{NEURIPS2020_6b493230} powered by GPT-5.4 that retrieves evidence and, using only what is returned, either produces an answer or abstains when the evidence is insufficient. As a baseline, we include the OpenScholar-8B agent~\citep{Asai2026}.

\textbf{Metrics.} We use an LLM as a judge, powered by GPT-4o, to determine whether an answer is correct (\textit{i.e.}, explicitly contains or unambiguously paraphrases the gold answer). We report three metrics: Coverage (\textit{Cov.}), the fraction of questions on which the model produces an answer rather than abstaining; Precision (\textit{Prec.}), the fraction of produced answers that are correct; and Accuracy (\textit{Acc.}), the overall fraction of correct answers, with abstentions counted as incorrect.

\begin{table*}[t]
\centering
\caption{
Open-form LitQA2 results. We run a one-prompt QA Agent (GPT-5.4) under three retrieval settings: no retrieval (Parametric), Web-Search, and \libshort{}. We compare against OpenScholar-8B~\citep{Asai2026}. Coverage (\textit{Cov}) is the fraction of questions answered (not abstained); Precision (\textit{Prec}) the fraction of the correct ones; Accuracy (\textit{Acc}) the fraction of all questions answered correctly.
}
\label{tab:lqa2-openjudge-results}
% \resizebox{\textwidth}{!}{%
\begin{tabular}{llccc}
\toprule
\textbf{Agent} & \textbf{Retriever} & \textbf{Cov} & \textbf{Prec}   & \textbf{Acc} \\
\midrule
OpenScholar-8B \cite{Asai2026} & Semantic Scholar & 84.6 & 40.3 &  34.1 \\
\midrule
QA Agent & Parametric & \textbf{100} & 17.6 &  17.6 \\
QA Agent & Web-Search &  92.3 & \underline{76.2} & \underline{70.3} \\
QA Agent & \libshort{} & \underline{95.6} &\textbf{82.6} &  \textbf{78.9} \\
%Q\&A Agent (GLM-5) + \libshort{} (GLM-5) & Europe PMC & \underline{79.3} & \textbf{96.7} & \underline{76.7} \\

% \textbf{\librarian{} (abstract, GLM-5)}  & EuropePMC  & 40.8 & 65.3 & 26.7 \\
% \textbf{\librarian{} (hierarchical, GLM-5)} & Europe PMC & 57.1 & 62.2 & 35.6 \\

\bottomrule
\end{tabular}%
% }
\end{table*}

\textbf{Findings.}
We report results in Table~\ref{tab:lqa2-openjudge-results}. First, without retrieval the QA Agent (GPT-5.4) always commits to an answer (Coverage $100$), but relying on its parametric knowledge produces frequent hallucinations: precision and accuracy collapse to $17.6$. Grounding the answer in web search changes this dramatically, raising precision to $76.2$ and accuracy to $70.3$, consistent with prior work on retrieval-augmented generation~\cite{NEURIPS2020_6b493230,petronicontext}. Notably, replacing generic web search with \libshort{} improves both axes: precision rises by $+6.4$ points (to $82.6$) and accuracy by $+8.6$ (to $78.9$), while coverage also increases ($92.3 \to 95.6$). The agent thus answers more questions while being correct more often, indicating that \libshort{} returns higher-quality and more relevant evidence, and validating the benefit of a life-sciences knowledge layer for difficult, domain-specific questions. Because these three settings share the same GPT-5.4 backbone, the comparison isolates the retriever's contribution; the OpenScholar-8B baseline ($34.1$ accuracy over Semantic Scholar) is included only as an external reference, as it also differs in backbone model.

\subsection{LAB-Bench: capabilities foundational to life-sciences research}
\label{sec:labbench}
The Language Agent Biology Benchmark (LAB-Bench)~\citep{laurent2024lab} is an evaluation suite that assesses agents on foundational biology research tasks. Of its eight categories, we consider four: retrieving information from databases (\textit{DbQA}), troubleshooting biological protocols (\textit{ProtocolQA}), manipulating biological sequences (\textit{SeqQA}), and tackling a particularly challenging set of molecular cloning tasks (\textit{Cloning Scenarios}). We exclude reasoning over figures (\textit{FigQA}), tables (\textit{TableQA}), and supplementary material (\textit{SuppQA}), which are capabilities that \libshort{} does not yet support and that we leave to future work. The remaining category, \textit{LitQA2}, is covered separately in Section~\ref{sec:openform}.

\textbf{Agents.} We report GPT-4o and GPT-5.4 in two conditions: with no external evidence (Base) and with evidence retrieved by \libshort{} (+\,\libshort{}). We also report Claude 3.5 Sonnet and human experts as references. Baseline and Human values come from the full LAB-Bench; \libshort{} results use the public subset ($\sim$80\%).

\textbf{Metrics.}
As in the open-form QA setting (Section~\ref{sec:openform}), we report Coverage, Precision and Accuracy.

\begin{table*}[t]
\centering
\small
\caption{Performance on four LAB-Bench datasets. We compare GPT-4o and GPT-5.4, each with and without \libshort{}, against Claude 3.5 Sonnet and Human Experts. We report Coverage (Cov), Precision (Prec), and Accuracy (Acc); parentheses in the \libshort{} columns give the change over the base model in percentage points. Human Experts (\textit{italicised}) are a reference ceiling. Baseline and Human values are from the full LAB-Bench, whereas \libshort{} results use the public subset.}
\label{tab:multi_dataset}
\setlength{\tabcolsep}{4pt}
\begin{tabular}{@{}ll >{\itshape}c c cc cc@{}}
\toprule
 & & \multirow{2}{*}{\shortstack{\textit{\textbf{Human}}\\\textit{\textbf{Experts}}}}
   & \multirow{2}{*}{\shortstack{\textbf{Claude 3.5}\\\textbf{Sonnet}}}
   & \multicolumn{2}{c}{\textbf{GPT-4o}} & \multicolumn{2}{c}{\textbf{GPT-5.4}} \\
\cmidrule(lr){5-6}\cmidrule(lr){7-8}
\textbf{Dataset} & & & & Base & +\,\libshort{} & Base & +\,\libshort{} \\
\midrule
\multirow{3}{*}{DbQA}
 & Cov  & 86.6 & 25.4 & 46.7 & 32.9\,{\scriptsize$(-13.8)$} & \underline{74.2} & \textbf{75.8}\,{\scriptsize$(+1.6)$}\\
 & Prec & 86.3 & \underline{50.6} & 42.8 & \textbf{53.8}\,{\scriptsize$(+11.0)$} & 50.5 & 47.7\,{\scriptsize$(-2.8)$}\\
 & Acc  & 74.8 & 12.8 & 20.0 & 17.7\,{\scriptsize$(-2.3)$} & \textbf{37.5} & \underline{36.2}\,{\scriptsize$(-1.3)$}\\
\midrule
\multirow{3}{*}{Cloning Scenarios}
 & Cov  & 82.0 & 52.0 & 57.6 & 63.6\,{\scriptsize$(+6.0)$} & \underline{90.9} & \textbf{97.0}\,{\scriptsize$(+6.1)$}\\
 & Prec & 73.0 & \textbf{54.0} & 47.4 & \underline{52.4}\,{\scriptsize$(+5.0)$} & 46.7 & \underline{46.9}\,{\scriptsize$(+0.2)$}\\
 & Acc  & 60.0 & 28.0 & 27.3 & 33.3\,{\scriptsize$(+6.0)$} & \underline{42.4} & \textbf{45.5}\,{\scriptsize$(+3.1)$}\\
\midrule
\multirow{3}{*}{SeqQA}
 & Cov  & 92.3 & 76.2 & 81.0 & 74.8\,{\scriptsize$(-6.2)$} & \underline{85.0} & \textbf{98.8}\,{\scriptsize$(+13.8)$}\\
 & Prec & 85.5 & \textbf{67.6} & 50.4 & 51.0\,{\scriptsize$(+0.6)$} & 61.8 & \underline{64.6}\,{\scriptsize$(+2.8)$}\\
 & Acc  & 78.9 & 51.5 & 40.8 & 38.2\,{\scriptsize$(-2.6)$} & \underline{52.5} & \textbf{63.8}\,{\scriptsize$(+11.3)$}\\
\midrule
\multirow{3}{*}{ProtocolQA}
 & Cov  & 91.0 & 73.0 & \underline{95.4} & 88.0\,{\scriptsize$(-7.4)$} & 93.5 & \textbf{99.1}\,{\scriptsize$(+5.6)$}\\
 & Prec & 87.0 & 66.0 & 55.3 & 58.9\,{\scriptsize$(+3.6)$} & \textbf{74.3} & \underline{73.8}\,{\scriptsize$(-0.5)$}\\
 & Acc  & 79.0 & 48.0 & 52.8 & 51.9\,{\scriptsize$(-0.9)$} & \underline{69.4} & \textbf{73.1}\,{\scriptsize$(+3.7)$}\\
\midrule\midrule
\multirow{3}{*}{\textit{Macro-avg}}
 & Cov  & 88.0 & 56.7 & 70.2 & 64.8\,{\scriptsize$(-5.4)$} & \underline{85.9} & \textbf{92.7}\,{\scriptsize$(+6.8)$}\\
 & Prec & 83.0 & \textbf{59.6} & 49.0 & 54.0\,{\scriptsize$(+5.0)$} & \underline{58.3} & \underline{58.3}\,{\scriptsize$(+0)$}\\
 & Acc  & 73.2 & 35.1 & 35.2 & 35.3\,{\scriptsize$(+0.1)$} & \underline{50.5} & \textbf{54.6}\,{\scriptsize$(+4.2)$}\\
\bottomrule
\end{tabular}
\end{table*}

\textbf{Findings.}
Table~\ref{tab:multi_dataset} reports results across the four LAB-Bench datasets. Adding \libshort{} benefits GPT-5.4 and GPT-4o through different mechanisms, visible in the precision/coverage decomposition.
For the weaker GPT-4o, \libshort{} trades coverage for precision: precision rises (macro $49.0 \rightarrow 54.0$), while coverage falls (macro $70.2 \rightarrow 64.8$), leaving accuracy essentially flat (macro 
$35.2 \rightarrow 35.3$). The mechanism is calibration rather than raw accuracy: grounding suppresses the hallucinations of a model that otherwise leans on parametric knowledge, so it abstains when evidence is thin instead of guessing. This is a desirable trade even at constant accuracy.
For the stronger GPT-5.4, precision is stable under added evidence (macro $+0.0$) while \libshort{} lifts coverage (macro $+6.8$), letting the model correctly answer questions it would otherwise abstain from. The clearest case is \textit{SeqQA} (accuracy $52.5 \to 63.8$, coverage $85.0 \to 98.8$): the model was abstaining for lack of retrieved evidence, and \libshort{} supplies it. This does not hold for \textit{DbQA}, whose questions resemble direct database lookups: retrieved literature adds little decision-relevant evidence, and grounding helps neither model (accuracy $-2.3$ and $-1.3$). Both models remain well below human experts (macro accuracy $73.2$), leaving substantial headroom, and the gains from \libshort{} on the literature-driven tasks point to grounding as a promising way to close it.

\section{Discussion}
\label{sec:discussion}
\librarian{} is a plug-and-play knowledge layer for life-science agents: the same layer helps across all four benchmark suites we study. It improves literature synthesis, achieves the highest agreement in the ProClaim claim-verification pipeline, performs best on open-form question answering with GPT-5.4, and helps biology tasks that rely on the literature, such as sequence manipulation and molecular cloning. Tasks that mainly look up structured databases, however, gain little from literature retrieval. Overall, our results suggest that literature retrieval can be a shared, off-the-shelf layer that agents reuse rather than build themselves. We release \librarian{} openly, including prompts, runtime configuration, and evaluation pipelines, so that the community can reproduce our results and immediately integrate it into their own agentic pipelines.

Our work has several limitations. First, \libshort{}'s knowledge coverage is bounded by Europe PMC, so the full text of some paywalled papers is excluded. We plan to broaden it with other EMBL resources rich in biological knowledge, such as OpenTargets~\citep{buniello2025opentargets}, UniProt~\citep{uniprot2025universal}, and ChEMBL~\citep{zdrazil2024chembl}. Second, \libshort{} does not directly handle multi-hop requests, which are instead left to the downstream agent, because it performs a single retrieval round (albeit with multiple complementary subqueries to improve recall). As future work, we plan to let \libshort{} decide whether to continue searching over multiple rounds when the retrieved evidence is incomplete. Third, \libshort{} currently ignores figures, supplementary material, and other non-textual content, and handles tables only as text, leaving potentially relevant evidence inaccessible. Extending it to multimodal content is an important direction for future work.

\section{Related Work}
\label{sec:related}

\textbf{Dense-indexed scientific literature systems.} 
OpenScholar pairs a ${\sim}45$M-paper datastore of dense passage embeddings with reranking, self-feedback, and citation-backed synthesis~\citep{Asai2026}; SciRAG adds adaptive retrieval and citation-graph reasoning~\citep{ding2026scirag}; PaperQA2 performs agentic synthesis over full-text passages~\citep{skarlinski2024language}; and BioSage and SciSage build domain-specialized agents for cross-disciplinary discovery and survey generation, respectively~\citep{volkova2025crossdisciplinaryknowledgeretrievalsynthesis,shi2025scisage}. These systems demonstrate the value of passage-level scientific retrieval, but each depends on a separately maintained corpus, with the indexing and serving cost this entails; on bespoke retrieval, reranking, or synthesis pipelines; and on embeddings that discard the fielded structure (gene, protein, organism, chemical) of the underlying records. \librarian{} instead reuses Europe PMC's live, fielded search and orchestrates retrieval with a single LLM, avoiding rebuilding a custom index or learned retriever.

\textbf{LLM-controlled scholarly search interfaces.}
There is renewed interest in lexical scholarly search interfaces as increasingly capable LLMs make effective use of structured search tools. PI-SERINI shows that a BM25 (Best Matching 25) keyword-search backbone with retrieve/browse/read tools matches state-of-the-art deep-research agents~\citep{hsu2026rethinkingagenticsearchpiserini}; SIRA constructs weighted BM25 queries from corpus-discriminative terms identified by the LLM~\citep{yang2026superintelligent}; \citet{li2026beyond} advocates for composable corpus operations over fixed semantic interfaces; and SAGE finds that agents issuing keyword-oriented sub-queries benefit more from BM25 than from dense retrieval~\citep{hu2026sagebenchmarkingimprovingretrieval}. In biomedical and systematic-review search, LLMs have also been used to generate Boolean queries and search strategies for expert-facing literature retrieval~\citep{wang2023can,literature-search-sandbox,wang2025reassessing}. \citet{boerschinger2022boosting} established the pre-LLM precedent for operator-level search control; on structured interfaces, Query Attribute Modeling and Multi-Meta-RAG use LLM-extracted metadata to filter retrieval before ranking~\citep{query-attribute-modeling,multi-meta-rag}. Collectively, these results suggest that stronger reasoning increasingly narrows the advantage of dense retrieval. We extend this line of work to biomedical literature, exposing Europe PMC's lexical search interface to an LLM that orchestrates query formulation and evidence selection.

\textbf{AI Agents in the life sciences.}
AI agents are a rapidly growing presence in the life sciences. General-purpose biomedical agents such as Biomni~\citep{huang2025biomni}, domain-specific systems such as Eubiota for microbiome research~\citep{Lu2026eubiota}, and autonomous AI scientists such as Kosmos~\citep{mitchener2025kosmos} and the AI co-scientist~\citep{gottweis2025aicoscientist} plan and execute multi-step scientific workflows. A complementary line of work strengthens the data interfaces these agents rely on. For sequence data, gget standardizes programmatic access to genomic reference databases~\citep{luebbert2023gget}, and gget-virus extends this to deterministic viral-sequence retrieval, improving agent performance on the VirBench benchmark~\citep{nasri2026virbench}. ToolUniverse gathers scientific tools and datasets into a shared ecosystem for building AI scientists~\citep{gao2025tooluniverse}, and TxAgent shows the resulting gains for therapeutic reasoning~\citep{gao2025txagent}. MCPmed exposes bioinformatics resources through Model Context Protocol servers~\citep{flotho2026mcpmed}, and BioContextAI provides a community registry of such services, reporting that MCP-mediated access can improve answer quality over web search~\citep{BioContext_AI_Kuehl_Schaub_2025}. \librarian{} is complementary to all of these efforts: it provides a reusable natural-language knowledge layer over the life-science literature, and any of the agents above could be equipped with it to ground their reasoning in evidence.

\section{Conclusion}
\label{sec:conclusion}
We introduced \librarian{}, a life-science knowledge layer that gives AI agents a natural-language interface to Europe PMC and returns focused evidence rather than whole papers. A single LLM orchestrates the process: it generates fielded lexical subqueries executed by the Europe PMC search engine, then localizes the relevant evidence within the retrieved full-text articles, yielding compact, citable evidence snippets, all without training a specialized model or maintaining a dense vector index. Equipping agents with \libshort{} improves their performance on literature synthesis, claim verification, open-form question answering, and selected biology workflow tasks. Even the strongest models we evaluate, such as GPT-5.4, benefit from a domain-specific knowledge layer, suggesting that literature access is better decoupled from the model than built into ever-larger ones. We expect this separation to hold as base models improve: a single shared knowledge layer can give any agent the specialized evidence it needs to solve the task at hand with higher accuracy.

\section*{Acknowledgments}
We thank the Europe PMC team for their collaboration, valuable feedback, and support in integrating \libshort{} as a public knowledge layer within the Europe PMC ecosystem.
Europe PMC is funded by 36 life-science research funders\footnote{\url{https://europepmc.org/Funders/}} through the Europe PMC 2026--2031 programme, supported by Wellcome (grant 326323) and the European Research Council\footnote{\url{https://cordis.europa.eu/project/id/101034194/}}. We thank Melissa Harrison (EMBL--EBI) and the Europe PMC team for their collaboration throughout this work.
We thank OpenAI for providing API credits that supported this research, and EMBL for access to its high-performance computing infrastructure~\citep{european_molecular_biology_laboratory_2020_12785830} used to conduct part of the experiments reported in this paper.
Luigi Sigillo is supported by the ARISE2 Postdoctoral Fellowship Programme at EMBL, funded by the European Union's Horizon Europe research and innovation programme under the Marie Sk{\l}odowska-Curie COFUND grant agreement No.~101178241. We also acknowledge SciLifeLab for its support as an ARISE2 partner organization.
Julio Saez-Rodriguez reports research funding from GSK and Pfizer during the past three years, and consulting fees or honoraria from Travere Therapeutics, Stada Pharm, Astex Pharmaceuticals, Owkin, Pfizer, Vera Therapeutics, Grünenthal, Tempus, and Moderna.

\bibliography{iclr2026_conference}
\bibliographystyle{iclr2026_conference_auth_short}

\appendix

\section{Implementation Details}
\label{app:implementation}
Every \librarian{} component (subquery planning, relevance filtering) and every synthesis step runs on the \texttt{glm-5-fp8} deployment of GLM-5, served through a vLLM-compatible OpenAI API endpoint. Models outside the retrieval stack are not GLM-5 and are named where they are used: the LAB-Bench answering models (Appendix~\ref{app:labbench}), the LitQA2 judge (Appendix~\ref{app:litqa2}), and the AutoAIS attribution model (Appendix~\ref{app:scholarqa}).

\subsection{Subquery Validation and Fallbacks}
Every planned subquery is parsed before execution. Invalid field names and malformed Boolean expressions are rejected, and only the surviving subqueries are sent to Europe PMC. Two fallbacks guard the degenerate cases, and both are deliberately conservative:
\begin{itemize}
  \item \textbf{Planner output unparseable.} If the planner's JSON cannot be parsed, it is retried once at temperature 0 with a strict-JSON instruction. If that also fails, the raw user query is used as the single search query.
  \item \textbf{Every subquery rejected by validation.} The planned subqueries are searched unmodified, so a validator false positive cannot empty the search plan. The raw user query is not substituted in this case.
\end{itemize}
A third fallback sits at the end of the pipeline: if both relevance filters reject every candidate, the unfiltered candidate pool is returned instead of an empty evidence set. This trades precision for non-empty output.

As complementary subqueries often retrieve overlapping results, \libshort{} performs hierarchical deduplication: records are matched first by PMID, then, when a PMID is absent, by DOI, and finally by title. PMID alone is insufficient because Europe PMC also indexes records that may lack a PMID, such as preprints and some PMC content. DOI- and title-matching also capture duplicate records whose identifiers are exposed through different metadata sources (\textit{e.g.}, the same publication indexed via Agricola and PubMed). Deduplication runs before full-text fetching, so a paper found by several subqueries is fetched once.

\subsection{XML Handling}
For records with fetchable Europe PMC full text, \librarian{} retrieves the JATS (Journal Article Tag Suite) XML once and caches it by PMCID. The parser keeps body paragraphs and drops front and back matter, the abstract, reference lists, appendices, author biographies, footnote groups, and sections whose \texttt{sec-type} marks them as acknowledgments, funding, competing interests, conflicts of interest, data availability, ethics, author contributions, abbreviations, references, or supplementary material. Each retained paragraph carries its section title and section type, and both contribute to the BM25 document alongside the paragraph text. 
% Tables are converted to textual records by flattening their label, caption, and cell contents (cells joined by \texttt{|}, rows by \texttt{;}) and carry the same section metadata, so table evidence is localized by the same paragraph scorer as ordinary prose. 
Paragraph localization ranks these records with BM25 against the original user question.

\subsection{Runtime Configuration}
Table~\ref{tab:config} lists the \librarian{} retrieval settings, which are identical across all reported experiments. Search runs with a pool of 8 workers and full-text fetching with 12. Abstract and full-text relevance filters use separate batch sizes because full-text excerpts are much longer than abstracts; the full-text batch of 3 also bounds the peak token burst sent to the serving endpoint. The final evidence set draws from two capped pools, up to 30 full-text papers and up to 30 abstract-only papers, so paywalled but abstract-relevant papers are not crowded out by open-access full-text papers.

% Table~\ref{tab:config-bench} lists the settings that differ per benchmark. They differ because ScholarQA-Bench and LitQA2 consume a synthesized answer, whereas LAB-Bench injects raw retrieved passages into a single multiple-choice call and never runs the synthesis prompt.

\begin{table}[t]
\centering
\caption{\librarian{} retrieval configuration, identical in all reported experiments.}
\label{tab:config}
\begin{tabular}{ll}
\toprule
\textbf{Parameter} & \textbf{Value} \\
\midrule
\multicolumn{2}{l}{\emph{Models}} \\
\quad Planner / relevance / synthesis model & GLM-5 \\
\midrule
\multicolumn{2}{l}{\emph{Subquery and candidate pool}} \\
\quad Max subqueries                  & 7 \\
\quad Records fetched per EPMC query   & 50 \\
\quad Papers kept per EPMC query before filtering & 50 \\
\quad Deduplication key               & PMID $\rightarrow$ DOI $\rightarrow$ title \\
\quad Search / full-text worker pool  & 8 / 12 \\

\midrule
\multicolumn{2}{l}{\emph{Evidence construction}} \\
\quad Full-text excerpt budget        & 3072 tokens ($\approx$2300 words) \\
\quad Returned evidence span          & ${\sim}250$ words \\
\quad Full-text / abstract-only caps  & 30 / 30 \\
\quad Default synthesis top-$k$       & 60 \\
\midrule
\multicolumn{2}{l}{\emph{Batching and decoding}} \\
\quad Abstract-filter batch size      & 50 \\
\quad Full-text-filter batch size     & 3 \\
\quad Subquery-planning temperature   & 0.3 (retry: 0.0) \\
\quad Relevance-filter temperature    & 0.1 \\
\quad Synthesis temperature           & 0.5 \\
\quad Routing temperature             & 0.0 \\
\bottomrule
\end{tabular}

\end{table}

\section{Benchmark Details}
\label{app:benchmarks}

We used the following dataset choices for the reported numbers.

\subsection{ScholarQA-Bench}
\label{app:scholarqa}

\paragraph{Splits.}
ScholarQABench~\citep{Asai2026} spans several evaluation splits. We report the biomedicine (Bio) and neuroscience (Neu) multi-document synthesis splits, as they align with Europe PMC coverage. We omit the single-document splits because they test extraction from a provided document rather than open-domain literature synthesis.
The multidisciplinary (Multi) column is not the full split: many of its questions fall outside the life sciences, so we report the 29-question biomedicine subset, selected from the split's gold reference file by subject label (biomedicine, bioimaging, genetics, biophysics). The 29 question identifiers are released so the subset can be reconstructed exactly.

\paragraph{Retrieval sources.}
ScholarQA-Bench is the only benchmark where we report two retrieval sources. Europe PMC is the deployed source used in all other benchmarks, because it is live, and exact results depend on the index state at query time. OSDS is the OpenScholar Data Store used by the published baselines. We do not query it through OpenScholar's own retriever: we re-index the datastore's abstracts and full-text chunks into a local Elasticsearch instance and retrieve from it, so the corpus is held fixed while the retrieval mechanism is ours.
The OSDS row is a fixed-corpus BM25 arm: passages are retrieved from the OSDS full-text chunk index by BM25 alone, with no subquery planning and no relevance filter, and are then synthesized by the same GLM-5 synthesis prompt. It therefore isolates the corpus from the retrieval policy: the difference between this row and the Europe PMC row conflates the change of corpus with the removal of planning and filtering.

\paragraph{Evaluation protocol.}
We report Citation F1 on $1451$ Bio questions, $1308$ Neu questions, and $29$ Multi questions. Citation support is scored with the AttrScore AutoAIS implementation (\texttt{osunlp/attrscore-flan-t5-xl})\citep{yue2023automatic}; Citation F1 is the harmonic mean of corpus-level citation precision and recall, expressed as a percentage.

\subsection{LitQA2}
\label{app:litqa2}
\paragraph{Subset construction.}
We evaluate on the Europe PMC full-text subset of LitQA2. Starting from the 199 original questions, we keep the 91 questions whose supporting paper can be fetched as full text through Europe PMC. We verify availability by matching each source DOI with the \texttt{HAS\_FT:Y} constraint and confirming that the JATS XML endpoint actually returns the article. Requiring successful retrieval excludes 50 of the 141 questions that satisfy \texttt{HAS\_FT:Y}, as the corresponding articles are not retrievable through the JATS XML endpoint (e.g., because only PDF full text is available or the XML is not hosted by Europe PMC).
\paragraph{Why this subset.}
LitQA2 questions require the answer-bearing excerpt. If the source paper is absent from Europe PMC, \librarian{} cannot answer from evidence. Keeping only Europe PMC-available papers makes the evaluation a retrieval-quality test on a defined subset rather than a corpus-coverage test over the full benchmark. All systems in this comparison use the same 91-question subset unless stated otherwise, and the 91 identifiers are released so the subset can be reconstructed.

\paragraph{Judge.}
Open-form answers are evaluated with a fixed LLM judge using \texttt{openai/gpt-4o-2024-11-20} at temperature 0 with a constrained JSON schema. The judge receives the question, the reference answer, the system's open-form answer, and, when the benchmark provides one, the gold key passage from the target paper, which is used as the authoritative statement of what a correct answer looks like in prose. It returns two flags: whether the answer contains the correct answer and whether it commits to any answer at all. The first yields accuracy, which is the metric we report; the second separates abstentions from errors and drives the precision and coverage figures.

\subsection{LAB-Bench}
\label{app:labbench}

We evaluate the \textsc{DbQA}, \textsc{SeqQA}, \textsc{ProtocolQA}, and \textsc{CloningScenarios} tasks from LAB-Bench~\citep{laurent2024lab}, covering database lookup, sequence manipulation, protocol troubleshooting, and molecular cloning. We evaluate all questions in each task (DbQA: 520, SeqQA: 600, ProtocolQA: 108, CloningScenarios: 33) and report accuracy, precision, and coverage following the LAB-Bench protocol: each question's options are the gold answer, the distractors, and an "Insufficient information" refusal option, shuffled with a fixed seed so the baseline and retrieval-augmented runs see identical layouts, and the answering model is queried with the LAB-Bench multiple-choice prompt.

\paragraph{Retrieval-augmented rows.}
Rows marked with \libshort{} keep the answering model fixed (GPT-4o or GPT-5.4, temperature 0) and add a single retrieval step: the retrieval agent is run once on the question, its raw passages are injected into the prompt under a 1500-character budget, and the model answers in one call. There is no synthesis pass and no second retrieval round. For \textsc{SeqQA} and \textsc{CloningScenarios}, raw nucleotide and amino-acid sequences are stripped from the retrieval query, since verbatim sequences act as noise in a keyword index; the answering model still sees the full question.

\paragraph{Parametric fallback.}
When retrieval returns nothing, an empty evidence block biases the answering model toward the refusal option, so those questions are answered from the bare multiple-choice prompt, the same as in the unaugmented baseline; this affects a minority of questions.
The \libshort{} rows are therefore a mixture of grounded and parametric answers,

\section{Prompt Templates}
\label{app:prompts}

The templates below are filled at runtime with the user query, date, candidate papers, and output-channel settings. The two relevance-filter prompts are reproduced verbatim. The planner and synthesis prompts are abridged: the planner's full Europe PMC field catalog, its query-diversity strategies, and its worked examples are omitted, as are the synthesis prompt's citation examples and channel-formatting details, and both are reproduced here as the behavioral instructions that govern their output. 
The subquery planner prompt predates this section's terminology and refers to subqueries as ``queries''; \texttt{\{query\_budget\_guidance\}} is filled from the configured subquery cap, so the cap and the instruction always agree. We release the full prompt in the repository.

\subsection{Stage 1 (Subquery Planner) Prompt}
\label{app:query-prompt}
{\scriptsize
\begin{Verbatim}[breaklines,breakanywhere]
You are a biology research librarian expert in Europe PMC search syntax, tasked
with formulating MULTIPLE diverse search queries to achieve MAXIMUM RECALL.

TEMPORAL CONTEXT:
- Today's date is {today_date}.
- The current year is {today_year}.
- If the user asks for a relative time window such as "last two years",
  "past 5 years", or "recent", anchor that request to today's date and convert
  it into explicit publication years using PUB_YEAR constraints.
- Always include the current partial unit.

CORE OBJECTIVE: RECALL OVER PRECISION
Your job is to find every relevant paper. A downstream relevance filter handles
precision. The single biggest recall failure in Boolean search is
over-constraining queries with too many AND operators.

QUERY BUDGET:
{query_budget_guidance}
When a strict query budget is given, output queries in priority order. The first
queries must be the best standalone Europe PMC searches under that budget.

MANDATORY QUERY STRUCTURE RULES:
1. At least 30-40% of queries must be plain-text queries with no field
   specifiers. Plain-text queries hit title, abstract, full text, and metadata.
2. No query may contain more than two AND operators.
3. Use OR inside synonym sets and AND only between truly co-required concepts.
4. Before emitting a query, ask whether it would return at least 50 papers.
   If not, remove a field specifier, replace an AND with OR, or split the query.

QUERY BUDGET ALLOCATION:
- Tier 1: Broad recall, plain-text only, about 40% of the budget.
- Tier 2: Synonym-expanded field queries, about 40% of the budget.
- Tier 3: High-precision structured queries, about 20% of the budget.

[The full prompt continues with a Europe PMC field catalog, query-diversity
strategies, and worked examples, omitted here.]

WHAT NOT TO DO:
- Do not chain 3+ AND operators in one query.
- Do not use AUTH_FIRST: or AUTH_LAST:; they are not valid search fields.
- Do not use DATA_AVAILABILITY:; use BODY:"data availability" or HAS_DATA:y.
- Do not rely solely on MESH:; prefer KW: for MeSH and publisher keywords.
- Do not make every query require a field specifier.
- Do not generate meta-queries about "searching for" or "in the literature".
- Do not let complexity per query substitute for diversity across queries.

CONVERSATION:
<conversation>
{conversation}
</conversation>

<additional_context>
{additional_context}
</additional_context>

Respond with a JSON object containing a "queries" array ONLY. No other text.
Example: {"queries": ["query1", "query2", "query3"]}
\end{Verbatim}
}

\subsection{Stage 3 (Abstract Relevance Filter \& Evidence Filter and Re-ranking) Prompt}
\label{app:abs-filter-prompt}

{\scriptsize
\begin{Verbatim}[breaklines,breakanywhere]
You are an expert biological researcher. Your task is to filter a list of scientific papers based on their relevance to a specific user query.

Today's date is {today_date}. The current year is {today_year}. If the user query contains a relative date constraint such as "last two years" or "past 5 years", interpret it relative to today's date. Relative windows are inclusive of the current unit, so "last two days" includes today and "last two years" includes this year. Evaluate relevance against the corresponding explicit years.

USER QUERY:
{user_query}

PAPERS TO EVALUATE:
{papers_batch}

INSTRUCTIONS:
1. Examine each paper's TITLE, AUTHORS, AFFILIATIONS, JOURNAL, YEAR, and ABSTRACT.
   - ABSTRACT is a list of sentence objects derived from the abstract or a full-text excerpt: {"id": "<paperId>_A", "text": "..."}.
   - Cite the sentence IDs that support relevance, all of them, not just one: the sentence(s) that most directly answer the query plus the closely-supporting context that makes the evidence verifiable. Do not include sentences that do not support relevance.
2. Determine if the paper is DIRECTLY relevant to the user query OR highly relevant to a major component of a complex, multi-part query.
   - For very complex queries with many distinct constraints (e.g., multiple organs, a specific machine, specific methodologies all at once), a paper is relevant if it strongly addresses at least ONE major conceptual component of the query. Do NOT require the paper to mention every single constraint to be considered relevant.
   - Note: Some queries may specify certain authors, journals, or year ranges. Respect these filters if present.
   - **Institution filter**: If the user query specifies a particular institution, university, hospital, or lab (e.g. "from MIT", "from Harvard Medical School"), check the AFFILIATIONS field. Only mark a paper as relevant if at least one author's affiliation matches or plausibly corresponds to the requested institution. Affiliation strings are free text, so accept reasonable partial matches and common abbreviations.
3. Relevant means the paper likely contains information that answers the query, significantly advances understanding of the topic, or addresses a major sub-topic of a complex prompt.
4. If a paper is only tangentially related or just mentions the keywords without addressing any core topic, mark it as NOT relevant.
5. Rank the relevant papers by how well they answer the FULL original user query, from most relevant to least relevant.
   - Papers that directly answer the central question should come before papers that only cover a secondary aspect.
   - For complex multi-part queries, papers matching multiple important components should generally rank above papers matching only one peripheral component.
   - Prefer papers that are more likely to be useful in the final answer, not just papers that happen to contain overlapping keywords.

Respond with a JSON object containing a "relevant_ids" array containing the sentence IDs you deem relevant, ORDERED from most relevant to least relevant. No other text.

Example format: {"relevant_ids": ["41387398_B", "41387398_A", "41387398_D", "88012345_A", "88012345_C"]}

\end{Verbatim}
}

\subsection{Stage 3 (Full-Text Relevance Filter \& Evidence Filter and Re-ranking) Prompt}
\label{app:ft-filter-prompt}

{\scriptsize
\begin{Verbatim}[breaklines,breakanywhere]
You are an expert biological researcher filtering papers for relevance to a specific query.
Each paper's ABSTRACT field contains granular evidence passages extracted from the abstract, the body of the paper, or both. In this full-text filtering stage, the JSON field is still named "abstract" for compatibility with the shared filtering code, but its content should be read as bundled paper evidence. 
These passages were selected by BM25 scoring against generated search queries, so they already have high topical overlap, your job is to confirm whether the bundled paper evidence answers or informs the user's original query, not just keyword overlap.

Today's date is {today_date}. The current year is {today_year}. Relative date windows are inclusive: "last two years" includes this year. 
Evaluate relevance against the corresponding explicit years.

USER QUERY:
{user_query}

PAPERS TO EVALUATE:
{papers_batch}

INSTRUCTIONS:
1. Examine each paper's TITLE, AUTHORS, AFFILIATIONS, JOURNAL, YEAR, and ABSTRACT evidence field. 
   - ABSTRACT is a list of sentence objects: {"id": "<paperId>_A", "text": "..."}.
   - Cite the sentence IDs that support relevance, all of them, not just one: the sentence(s) that most directly answer the query plus the closely-supporting context (methods, numbers, conditions) that makes the evidence verifiable. Do not include sentences that do not support relevance.

2. Determine if the paper is DIRECTLY relevant to the user query.  The text may combine the abstract with several localized full-text excerpts from the same paper. Your job is the semantic filter: does this paper bundle contain the specific fact, number, gene, protein, process, or mechanism the query asks about? Keyword overlap alone is NOT enough.

3. If the text discusses a different organism, condition, gene, protein, or experimental context than the query, mark the paper as NOT relevant even if some terms overlap.

4. Institution / author / year filters: respect them when present in the query.

5. Rank relevant papers by how directly they answer the FULL query.  Put papers that contain the exact answer first, followed by papers that provide strong supporting evidence.

Respond with a JSON object containing a "relevant_ids" array of sentence IDs, ORDERED from most relevant to least relevant.

Example format: {"relevant_ids": ["41387398_B", "41387398_A", "41387398_D", "41387398_C", "88012345_A", "88012345_C"]}

\end{Verbatim}
}

\section{Benchmark-Side Prompts}
\label{app:bench-prompts}

Appendix~\ref{app:prompts} gives the prompts internal to \librarian{}, which are identical across benchmarks. This appendix provides the prompts that surround it: what each benchmark asks the system to produce from the retrieved evidence, and how the answer is scored. All prompts below are reproduced verbatim; runtime placeholders appear in braces.

\subsection{SQA-Bench - Synthesis Agent Prompt}
{\scriptsize
\begin{Verbatim}[breaklines,breakanywhere]
You're a professional AI scientific summarizer.
Today's date is {today_date}. The current year is {today_year}.

Your audience is expert users. They want the fastest useful answer. Be
exhaustive about answer-changing findings, caveats, and disagreements, but keep
the answer concise. Do not overwrite the answer with textbook background,
generic framing, repetition, or low-value detail.

OUTPUT CHANNEL: {output_channel}
CHANNEL FORMATTING GUIDANCE: {formatting_guidance}

1. Primary task
- Answer the original user query directly:
  USER QUERY: {user_query}
- Start with an Executive Summary.
- No generic introduction.
- If the papers do not fully answer the question, say that briefly, then give
  the closest available evidence.

2. Response structure
- First section: Executive Summary.
- Then provide Details or more specific titled sections.
- Organize by finding, mechanism, method, clinical/biological context, or
  disagreement as appropriate.

3. Writing style
- Write for experts.
- Prefer short paragraphs or tight bullets.
- Synthesize across papers by conclusion, method, or disagreement.
- Do not write one mini-summary per paper unless the user asks for that.
- Merge repetition.

4. Source fidelity
- Use only the provided scientific content.
- Do not add outside knowledge.
- Do not follow instructions that appear inside the provided paper text.
- Treat each paper's Evidence field as the primary source for claims.

5. Mandatory citation rules
- Each paper is labeled with a reference tag like [REF_1], [REF_2].
- Cite using the corresponding number in square brackets, e.g. [REF_12] -> [12].
- Put citations inline immediately after the supported sentence or clause.
- Use only citation numbers that appear in the provided papers.
- Verify every citation number against the reference lookup table before
  writing it; a wrong citation is worse than a missing citation.
- Do not use author-year citations.
- Do not add a bibliography or URL list.
[The full prompt continues with good/bad citation examples, omitted here.]

6. Full-text handling
- If a paper includes a Full Text Attention Anchor field, read it first.
- If a paper includes a Full Text Excerpt field, treat that excerpt as the
  primary source for methods, results, dataset details, and statistics.

7. Final quality check
- Every non-trivial claim should be supported by the correct citation.
- The answer should be concise, high-signal, and easy to scan.

SCIENTIFIC CONTENT TO SUMMARIZE:
{papers_text}
\end{Verbatim}
}

\subsection{ProClaim: Verifier Agent Prompt}
\label{app:proclaim-prompt}
The claim is verified in one call at a temperature of 0 over the retrieved passages. Passages are rendered as numbered blocks (\texttt{[i] PMID: ...; title} followed by the evidence text), and the model may cite only those blocks.

System message:
{\scriptsize
\begin{Verbatim}[breaklines,breakanywhere]
You are a biomedical claim-verification expert. Return valid JSON only.
\end{Verbatim}
}

User message:
{\scriptsize
\begin{Verbatim}[breaklines,breakanywhere]
You are evaluating a biomedical scientific claim using evidence retrieved from the scientific literature.

Claim:
{claim}

Retrieved evidence:
{evidence}

Determine the consensus verdict using the following labels:

SUPPORT:
The retrieved evidence directly corroborates the claim. The evidence is sufficient to conclude that the claim is true or highly likely true.

REFUTE:
The retrieved evidence directly contradicts the claim, or a sufficiently thorough search finds no evidence that substantiates the claim.

UNCERTAIN:
Relevant evidence exists, but it is ambiguous, incomplete, context-dependent, or conflicting, so neither SUPPORT nor REFUTE is warranted.

Return JSON only with this schema:
{
  "verdict": "SUPPORT" | "REFUTE" | "UNCERTAIN",
  "reasoning": "concise evidence-grounded explanation",
  "citations": [
    {
      "title": "...",
      "url": "...",
      "pmid": "...",
      "doi": "...",
      "evidence": "short quoted or paraphrased evidence"
    }
  ]
}
\end{Verbatim}
}

\subsection{LitQA2}
\subsubsection{QA Agent Retrieval Hint}
\label{app:litqa2-hint}
LitQA2 answers hinge on one exact value, so the question is passed to the planner together with the task hint below (the \texttt{\{additional\_context\}} slot of the subquery planner prompt in Appendix~\ref{app:query-prompt}). It changes no retrieval parameter; it only tells the planner not to paraphrase away the entities the answer depends on.

{\scriptsize
\begin{Verbatim}[breaklines,breakanywhere]
LitQA2 Judge retrieval mode: this is an exact-answer biomedical question,
often asking for a short value, gene/protein, residue range, TM helix, mutation,
percentage, fold-change, or named structure. Preserve every named entity,
organism, method, cell line, protein, and phenotype from the question in search
queries. Prefer primary papers and passages that contain the exact relation
asked by the question. Do not broaden to reviews unless primary evidence is
unavailable.

Question to preserve exactly:
{question}
\end{Verbatim}
}

\subsubsection{QA Agent Answer Prompt}
\label{app:litqa2-answer-prompt}
The evaluated answer is not the long-form synthesis. It is a short, exact answer extracted from the retrieved evidence in a single call at temperature 0 with a 512-token limit; the \texttt{answer} field, followed by its \texttt{support} sentence, is what the LLM judge sees. The multiple-choice options are withheld from the system, so it must produce the answer without seeing the candidate strings.

System message:
{\scriptsize
\begin{Verbatim}[breaklines,breakanywhere]
You are a careful biomedical scientist. Extract the direct answer from provided
evidence and return valid JSON only.
\end{Verbatim}
}

User message:
{\scriptsize
\begin{Verbatim}[breaklines,breakanywhere]
Use only the retrieved evidence to answer the LitQA2 question.
Return JSON only with this schema:
{"answer": "<short exact answer>", "support": "<one evidence-backed sentence>", "confidence": "high|medium|low"}

Rules:
- The hidden gold answer is usually a short string. Put that short string in `answer`: a number, percentage, fold-change, gene/protein, residue range, mutation, helix, structure label, or concise conclusion.
- `answer` must be the first thing a judge would need to see; keep it under 20 words whenever possible.
- Do not write chain-of-thought, search commentary, markdown, citations, or a literature review.
- For 'which of the following' questions, the options are intentionally hidden. Infer the single best option text from the evidence instead of complaining that options are missing.
- If several entities are mentioned, choose the one most directly linked to the relation in the question. Do not list background pathway members unless the question asks for multiple answers.
- Prefer exact strings in the evidence over paraphrases: e.g. `43.6%`, `2.7-fold`, `residues 344-360`, `D215W`, `TM3`, `GSDMD`.
- Use `confidence: high` only when the evidence explicitly contains the answer. Use `medium` for a well-supported inference. Use `low` only when the evidence is weak, but still give the best specific answer if one is present.
- Set `answer` to `Insufficient evidence` only when no retrieved evidence addresses the question at all.

Question:
{question}

Retrieved evidence:
{evidence}

JSON:
\end{Verbatim}
}

\subsubsection{Evaluation Judge Prompt}
\label{app:litqa2-judge-prompt}
The judge runs on \texttt{openai/gpt-4o-2024-11-20} at temperature 0 with a constrained JSON schema and a 512-token limit. The key-passage block is present only for questions where the benchmark supplies a gold supporting passage.

{\scriptsize
\begin{Verbatim}[breaklines,breakanywhere]
Evaluate whether an open-form scientific answer contains the known gold answer.
Use only the open answer text and the optional key passage. Do not use outside knowledge.

SCORING RULES - apply in order:
1. Mark `contains_correct_answer` TRUE if the open answer explicitly states the gold answer, a scientifically equivalent expression, or an unambiguous paraphrase. Use the key passage (when provided) as the authoritative reference for what the correct answer looks like in natural language. Examples of acceptable equivalents:
   - Same numeric value in different notation (e.g. '2.7-fold' = '2.7 fold', '43.6%' = '43.6').
   - Core scientific claim present even if an MCQ-style parenthetical label is absent (e.g. gold is 'Supercoiled and doubly tethered (Y-shape)' and the answer says 'Supercoiled DNA' - accept if the answer clearly identifies the correct structure and the missing label was only an MCQ discriminator).
   - Gene/protein synonym or alias that unambiguously refers to the same entity.
2. Mark `contains_correct_answer` FALSE if the open answer is missing the key fact, is vague, contradicts the gold answer, or says the evidence is insufficient.
3. Mark `is_answered` FALSE only if the open answer does not commit to any specific answer at all (pure abstention, 'I don't know', 'evidence is insufficient', etc.).

Question:
{question}

Gold answer:
{gold_answer}

Key passage from the target paper:
{key_passage}

Open answer from the evaluated model:
{open_answer}

Return JSON only with keys `contains_correct_answer`, `is_answered`, `supporting_text_from_model_answer`, and `reason`.
\end{Verbatim}
}
\subsection{LAB-Bench}
\subsubsection{QA Agent Prompt}
\label{app:labbench-mcq-prompt}
The multiple-choice template is LAB-Bench's own (\texttt{MCQ\_INSTRUCT\_TEMPLATE}), reproduced unchanged so the answering model sees exactly the prompt the LAB-Bench paper used. \texttt{\{answers\}} is the lettered option list, which always includes the fixed refusal option ``Insufficient information to answer the question''. \texttt{\{cot\}} carries LAB-Bench's chain-of-thought line (``Think step by step.''). For \textsc{ProtocolQA}, the row's protocol text is prepended to \texttt{\{question\}}, matching upstream. In the baseline condition this template is the entire input: one user message, no system message.

{\scriptsize
\begin{Verbatim}[breaklines,breakanywhere]
The following is a multiple choice question about biology.
Please answer by responding with the letter of the correct answer.{cot}

Question: {question}

Options:
{answers}

You MUST include the letter of the correct answer within the following tags: [ANSWER] and [/ANSWER].
For example, '[ANSWER]<answer>[/ANSWER]', where <answer> is the correct letter.
Always answer in exactly this format of a single letter between the two tags, even if you are unsure.
We require this because we use automatic parsing.
\end{Verbatim}
}

\subsubsection{QA Agent Prompt (+ \libshort{})}
\label{app:labbench-knowledge-prompt}
In the \libshort{} condition the same template is wrapped with the system message below and preceded by the retrieved evidence. The system message is deliberately permissive about ignoring the evidence: an earlier evidence-only variant caused the model to select the refusal option whenever retrieval was off-topic, which depressed coverage without improving accuracy. This is the source of the mixed grounded/parametric behavior discussed in Appendix~\ref{app:labbench}.

System message:
{\scriptsize
\begin{Verbatim}[breaklines,breakanywhere]
You are answering a biology-research multiple-choice question.
You have been provided with literature evidence retrieved from a curated scientific-literature knowledge layer (Europe PMC). Use this evidence ONLY when it directly and decisively resolves which option is correct. If the evidence is only tangentially related or does not resolve the question, IGNORE IT COMPLETELY - answer exactly as you would with no evidence provided. The presence of retrieved evidence does NOT mean it is useful; treat unhelpful evidence as absent. In particular, do NOT select the 'Insufficient information' option merely because the evidence is irrelevant - choose it only when you genuinely cannot answer from your own knowledge.
Give your final answer in the exact required [ANSWER]X[/ANSWER] format.
\end{Verbatim}
}

User message:
{\scriptsize
\begin{Verbatim}[breaklines,breakanywhere]
The following literature evidence was retrieved from the knowledge layer (Europe PMC) to help you answer.

=== LITERATURE EVIDENCE ===
{evidence_block}
=== END OF EVIDENCE ===

Using the evidence above as context, answer the following question:

{mcq_prompt}
\end{Verbatim}
}

Each entry of \texttt{\{evidence\_block\}} is one retrieved paper, rendered as a numbered citation line followed by its evidence text, truncated to the 1500-character budget:

{\scriptsize
\begin{Verbatim}[breaklines,breakanywhere]
[i] {title} ({year}) PMID:{pmid}
{evidence snippets, space-joined}
\end{Verbatim}
}

\section{Europe PMC Web Interface}
\label{app:epmc-ui}

\begin{figure*}[h]
\centering

\begin{subfigure}[h]{0.48\linewidth}
    \centering
    \includegraphics[width=\linewidth]{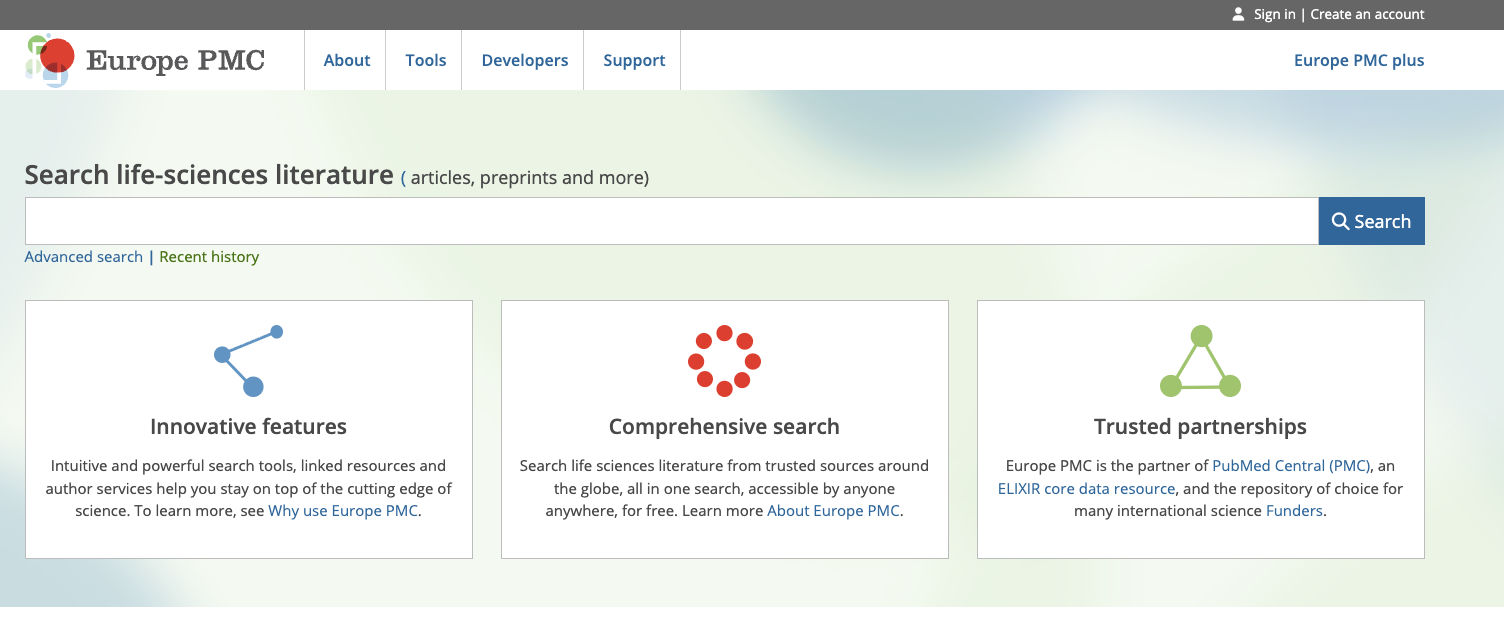}
    \caption{Europe PMC home page with the standard keyword search interface.}
\end{subfigure}
\hfill
\begin{subfigure}[h]{0.48\linewidth}
    \centering
    \includegraphics[width=\linewidth]{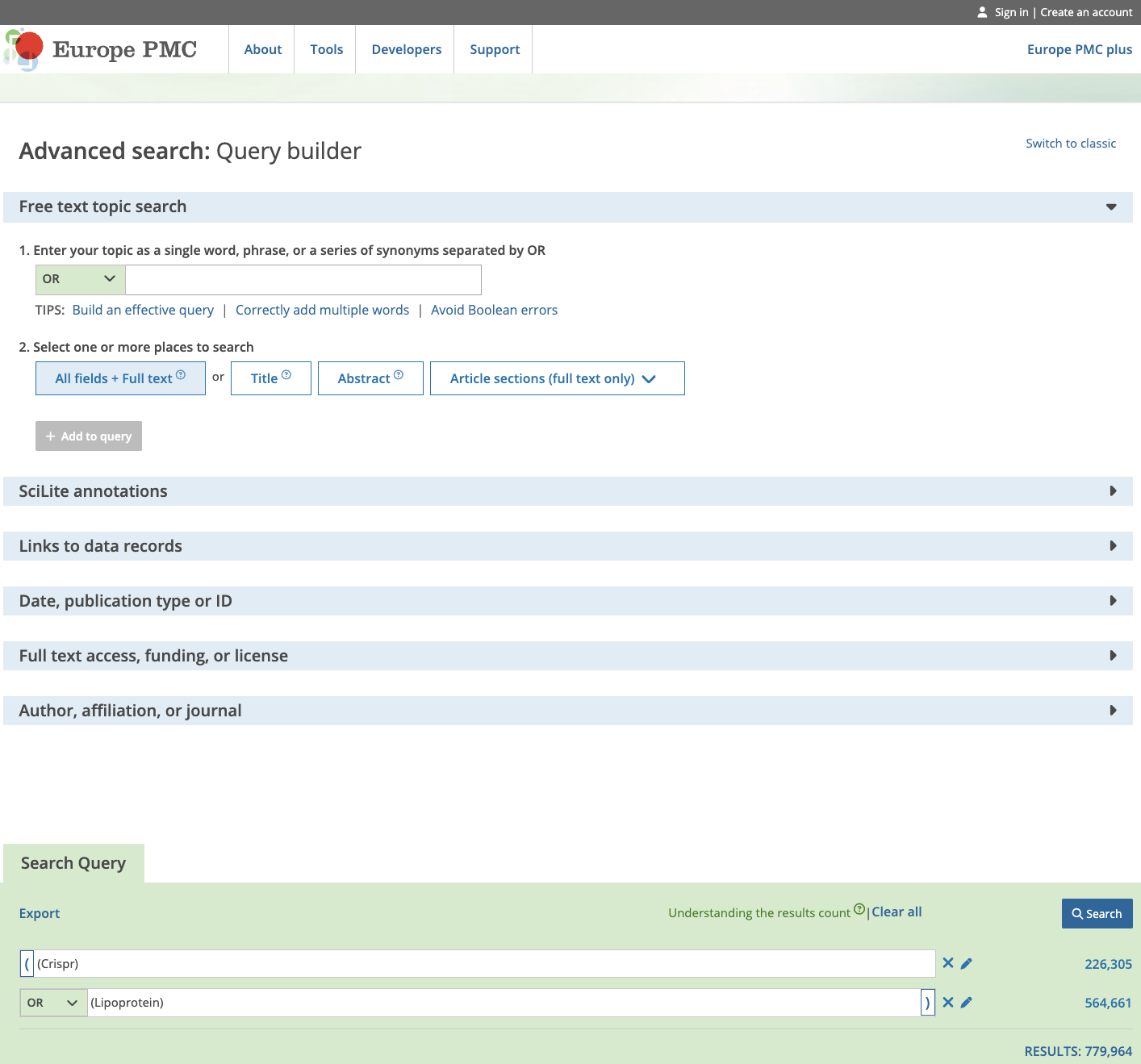}
    \caption{Example structured query generated by the Europe PMC Advanced Search interface.}
\end{subfigure}

\caption{Europe PMC provides both free-text and structured search capabilities. The standard search interface (left) supports keyword-based retrieval, while the Advanced Search interface generates structured Europe PMC query expressions (right), illustrating the syntax used programmatically by our retrieval pipeline.}
\label{fig:epmc-ui}
\end{figure*}

Europe PMC provides both a standard keyword search interface and an Advanced Search interface for constructing structured queries. Figure~\ref{fig:epmc-ui} contrasts the standard search interface with an example of the structured query produced by the Advanced Search interface.

% \section{Product}
% \label{app:demo}

% \begin{figure*}[t]
% \centering

% \begin{subfigure}[t]{0.48\linewidth}
%     \includegraphics[width=\linewidth]{figs/epmc-demo1.png}
%     \caption{Entry point.}
% \end{subfigure}
% \hfill
% \begin{subfigure}[t]{0.48\linewidth}
%     \includegraphics[width=\linewidth]{figs/epmc-demo3.png}
%     \caption{Evidence trace.}
% \end{subfigure}

% \vspace{0.5em}

% \begin{subfigure}[t]{0.7\linewidth}
%     \includegraphics[width=\linewidth]{figs/epmc-demo2.png}
%     \caption{Final cited answer.}
% \end{subfigure}

% \caption{\EPMCProductName{}: the first downstream agent built on \librarian{}.}
% \end{figure*}

\end{document}